\documentclass[letterpaper, 10pt, conference]{IEEEtran}

\IEEEoverridecommandlockouts                              

\usepackage{graphics} 
\usepackage{epsfig} 
\usepackage{mathptmx} 
\usepackage{times} 
\usepackage{amsmath} 
\usepackage{amssymb}  
\usepackage{caption}
\usepackage{subcaption} 
\usepackage{xcolor} 
\usepackage{numprint} 

\title{\LARGE \bf
Implementation of Road Safety Perception in 
Autonomous Vehicles in a Lane Change Scenario
}

\author{Enrico del Re \emph{Student Member, IEEE}, 
Cristina Olaverri-Monreal \emph{Senior Member, IEEE}
\thanks{Johannes Kepler University Linz, Austria; ITS-Chair  Sustainable Transport Logistics 4.0. \texttt{\{enrico.del\_re, cristina.olaverri-monreal\}@jku.at}}}

\begin{document}

\maketitle


\begin{abstract}

Understanding human driving behavior is crucial to develop autonomous vehicles' algorithms. However, most low level automation, such as the one in advanced driving assistance systems (ADAS), is based on objective safety measures, which are not always aligned with what the drivers perceive as safe and their correspondent driving behavior. Finding the bridge between the subjective perception and objective safety measures has been analyzed in this paper focusing specifically on lane-change scenarios. Results showed statistically significant differences between what is perceived as safe by drivers and objective metrics depending on the specific maneuver and location of drivers. 

\end{abstract}

\section{INTRODUCTION}
Augmenting road safety is among others one of the objectives of autonomous vehicles (AVs). This is due to the fact that more than 90\% of traffic accidents can be traced back to human errors and replacing human drivers by automation is expected to cause fewer traffic-related injuries \cite{olaverri2016autonomous}, (cited in~\cite{olaverri2020promoting}). 
Driver's inattention, as well as decision errors that are for example the consequence of a wrong traffic environment perception can jeopardize safety.

In this context, the subjective perception of the possibility of harm may relate to the risk of being involved in a traffic accident. 
Factors related to the state of the driver that might augment the probability of a road risk can be measured by comparing them with standard physiological and workload metrics \cite{subjective}.
An objective assessment of traffic safety on the other hand can be achieved by using vehicle-related parameters such as distance to other vehicles in the vicinity.

We address safety in this paper by focusing on different metrics obtained from a lane-change maneuver after performing an overtaking maneuver, in a highway scenario. We argue that a situation that a driver perceives as safe (subjective safety, expressed in the road by a distance to the vehicle ahead or behind) varies depending on whether the vehicle being driven is a leading (LV) or a following vehicle (FV).

Under the assumption that lane-change maneuvers are only being performed if the driver assesses the situation as safe, it is imperative to investigate traffic situations after an overtaking maneuver has been performed and lane-changes can occur safely for two isolated vehicles.
Further we compare those results with lane-changes that occur in other road situations to find out if a different applicable safety index should be part of AV algorithms.
As autonomous vehicles are expected to display human-like behavior, at least to the extent that their decisions are being understood by other road users~\cite{smirnov2021game}, such an index could also reflect what human drivers perceive as safe.

A related topic is driving behavior classification in safe, normal and aggressive, depending on acceleration and speed parameters ~\cite{validi2022metamodel}. In this sense, a ``courteous'' behavior from the side of the AV could contribute to be perceived as more human-like~\cite{Courteous}.

The remaining parts of this paper are organized as follows: the following section describes related work. Section~\ref{sec:method} describes the applied methodology including the studied lane-change scenarios, the dataset utilized for this study, the description of the safety measures, and the applied statistical analysis. 
The results from the analysis are presented in section~\ref{sec:results}; and the section~\ref{sec:conclusion} concludes the work and outlines future research.

\section{Related Work}
\label{sec:RelatedWork}

Both subjective and objective traffic safety have been heavily researched topics, though mostly separately.
Surveys connecting subjective safety with accident reports \cite{factorslanechange}, road conditions and driving behavior \cite{ESRA} differ strongly from most papers regarding objective safety measures such as \cite{SSM}, both in scope and methodology.

Subjective safety analysis has been typically performed focusing on surveys that described the type of behavior or infrastructure that affected traffic accidents. Not wearing a seat-belt, distracted driving behavior such as using a phone, driving over the speed limit, but also road conditions, traffic signals and low road visibility are common factors that were mentioned to be addressed by policy makers. 

On the other hand, objective safety measures based on standard metrics are used objectively to evaluate the safety of infrastructure or ADAS.

Fewer research articles have combined both safety aspects. For example, the authors in \cite{subjective} combined individual subjective safety with large-scale objective safety parameters while the authors in \cite{DSSM} compared acceleration and deceleration of vehicles, analyzing indirectly metrics that resulted from a perceived, subjective safety estimation, to be compared with objective safety metrics that indicated a minimum risk. For example, an ADAS based on Time To Collision (TTC), a frequently used objective-safety measure, is supposed to decelerate earlier and more frequently in the case its velocity exceeds the velocity of the vehicle ahead.
The observed human behavior differed in many cases from the acceleration and deceleration patterns that were implemented as standard measures in ADAS.

In the related literature objective safety has been investigated without consideration of driver perception characteristics that can be due to different vehicle positions in traffic lanes. We contribute in this paper to the body of knowledge by addressing lane changes maneuvers after an overtaking in order to find out if a safety index should be incorporated to autonomous vehicles algorithms. To the best of our knowledge such an approach has not been followed in previous research.  

\section{Applied Methodology}
\label{sec:method}

\subsection{Lane change scenario}

We selected a scenario that consisted of a lane-change performed after an overtaking situation and extracted the independent variables that affect safety.

\begin{figure}
     \centering
     \begin{subfigure}[b]{0.45\linewidth}
         \centering
         \includegraphics[width=\textwidth]{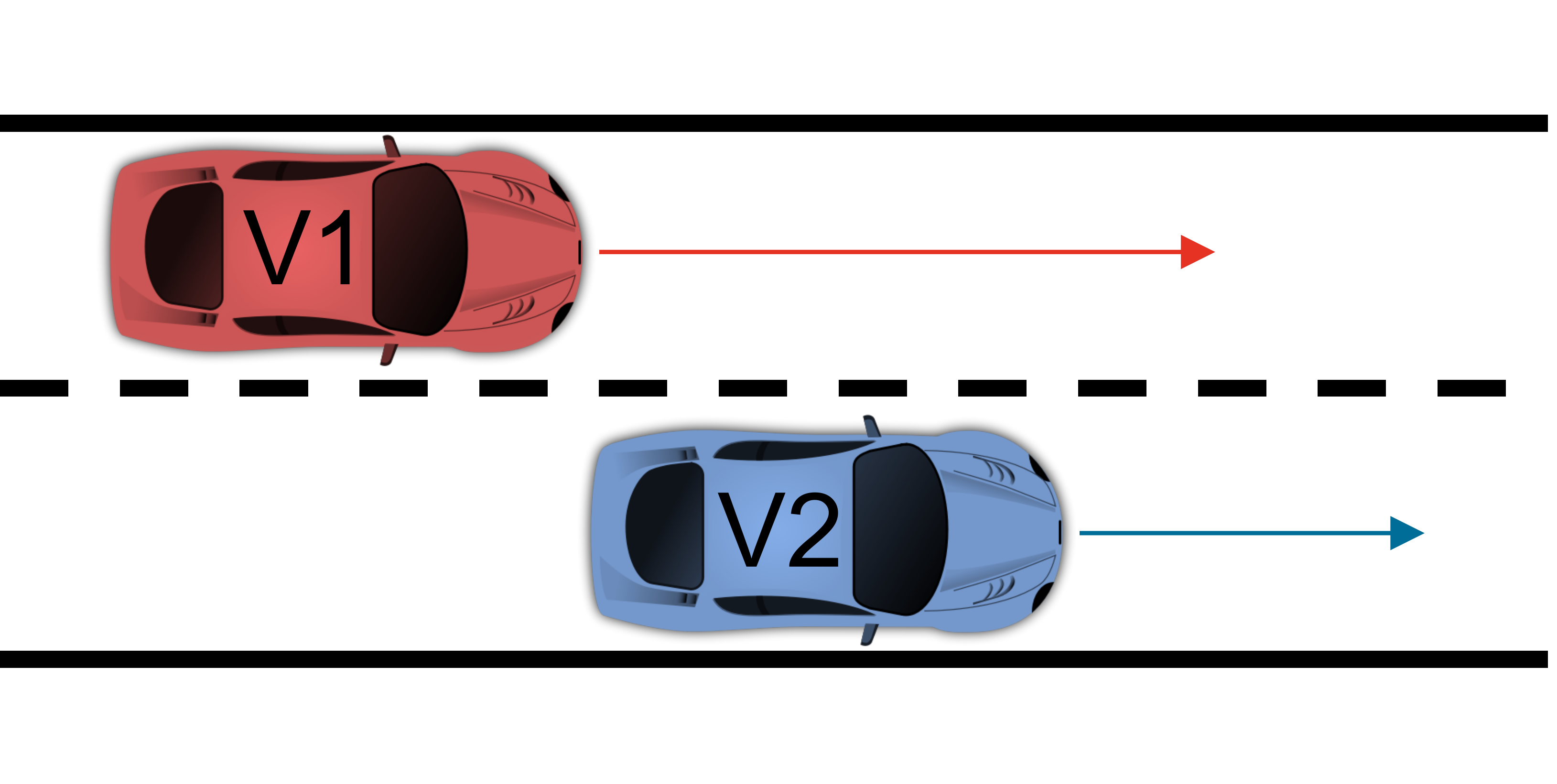}
         \caption{Beginning}
         \label{fig:ov_type1}
     \end{subfigure}
     \hfill
     \begin{subfigure}[b]{0.45\linewidth}
         \centering
         \includegraphics[width=\textwidth]{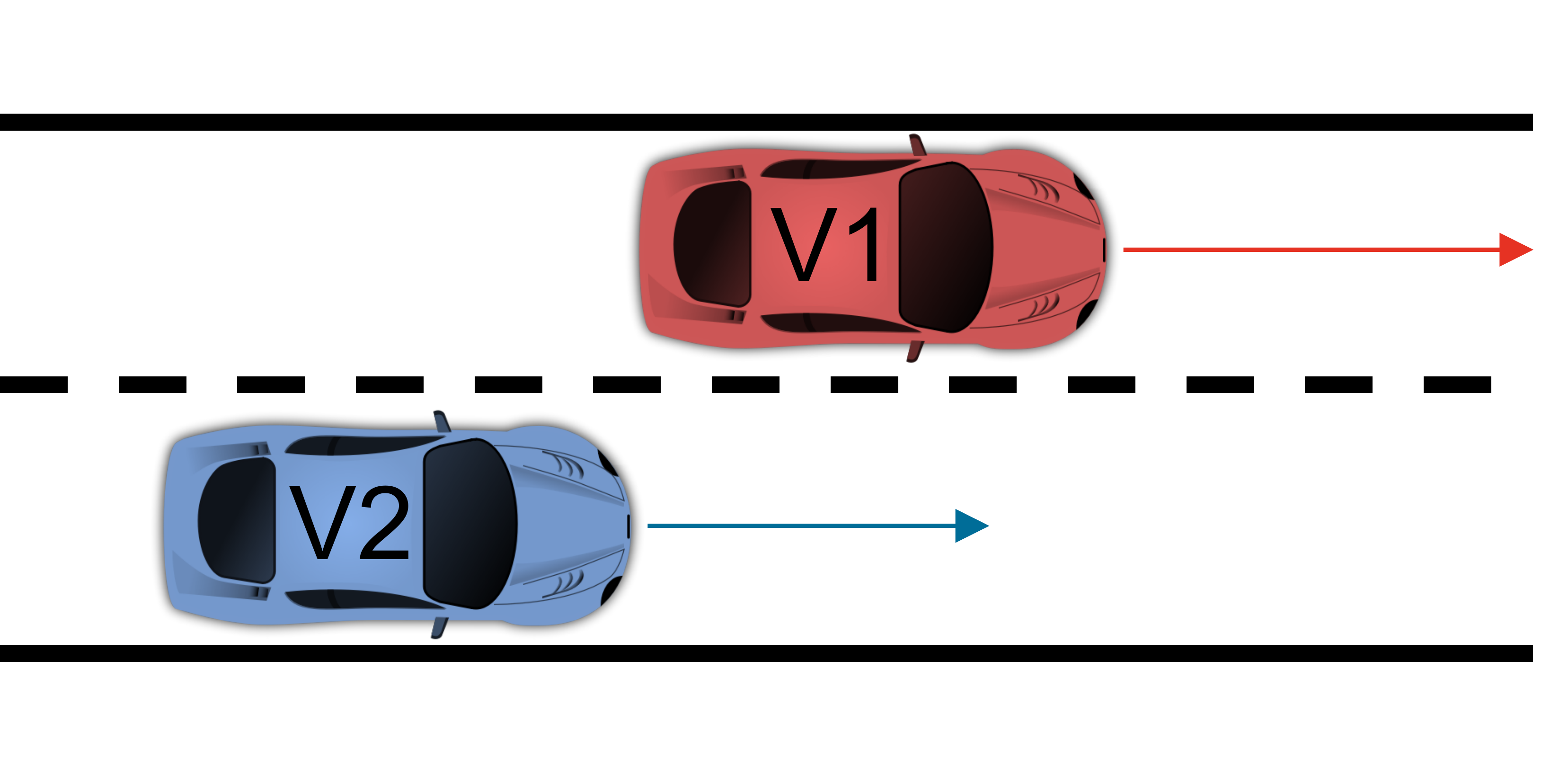}
         \caption{End}
         \label{fig:ov_type1P}
     \end{subfigure}
        \caption{Overtaking as defined for this paper. The vehicle V1 (red) passes the vehicle V2 (blue) on the left adjacent lane.}
        \label{fig:overtaking}
\end{figure}

Overtaking in this paper is defined by a vehicle (V1) with relatively higher speed surpassing another vehicle (V2) by using the left adjacent lane. The maneuver is visualized in Figure \ref{fig:overtaking}. We only considered interaction between cars with similar sizes. The exact size of the vehicle should not noticeably influence the scenario unless it implies a different driving behavior, such as the different stopping distance of a truck.
Following the overtaking maneuver the following two lane-changes are possible:\\

\begin{figure}
     \centering
     \begin{subfigure}[b]{0.45\linewidth}
         \centering
         \includegraphics[width=\textwidth]{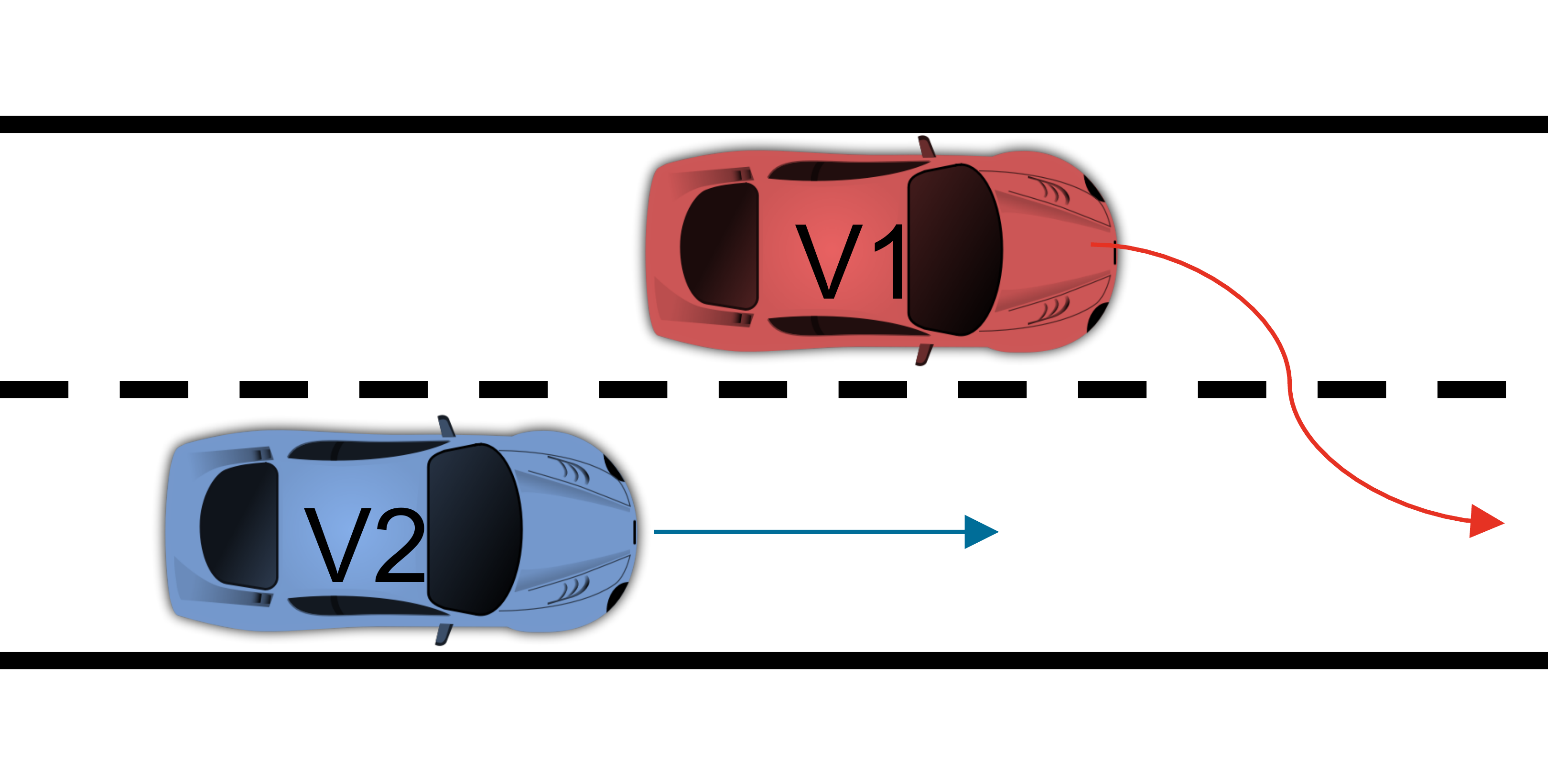}
         \caption{Beginning}
         \label{fig:lcA1}
     \end{subfigure}
     \hfill
     \begin{subfigure}[b]{0.45\linewidth}
         \centering
         \includegraphics[width=\textwidth]{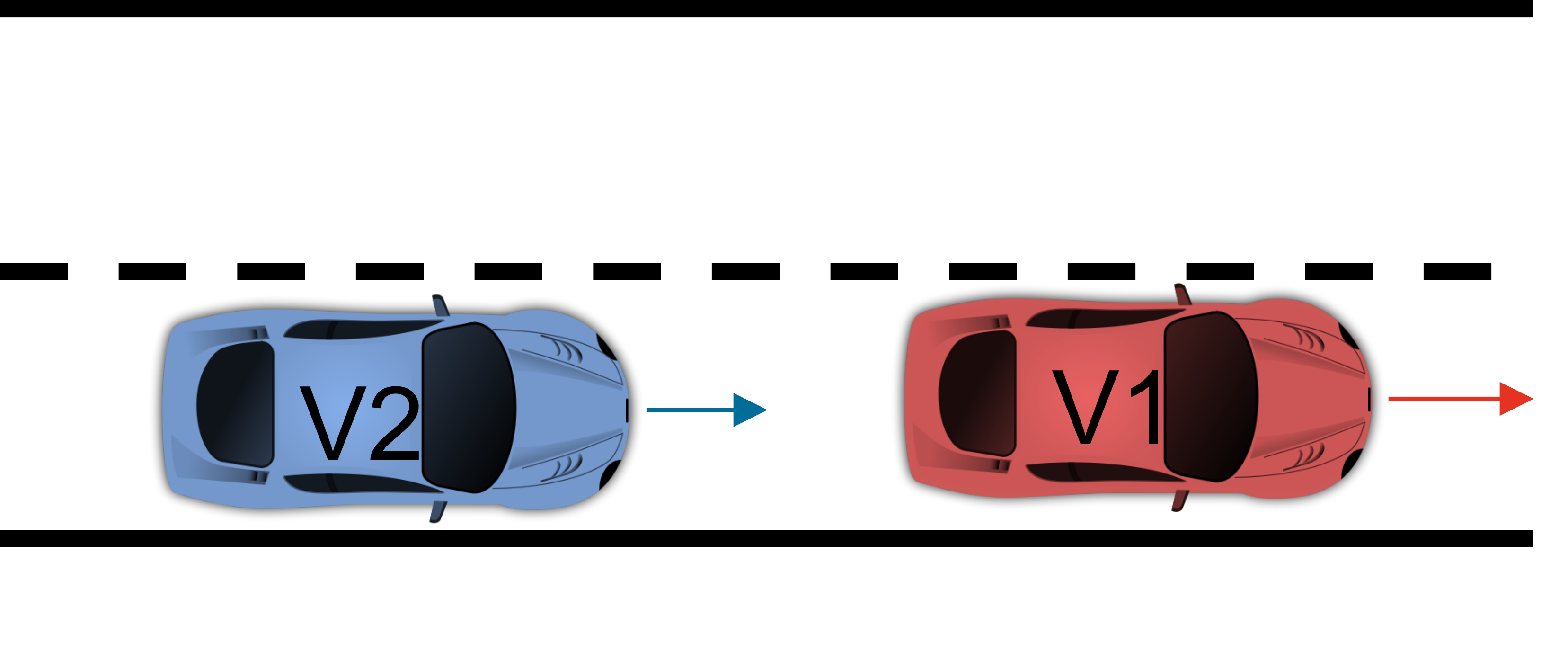}
         \caption{End}
         \label{fig:lcA2}
     \end{subfigure}
     \caption{1st type of lane change. The vehicle V1 (red) changes lane in front of the vehicle V2 (blue) to the right adjacent lane.}
     \label{fig:lc_typeA}
\end{figure}

\begin{figure}
	\centering
     \begin{subfigure}[b]{0.45\linewidth}
         \centering
         \includegraphics[width=\textwidth]{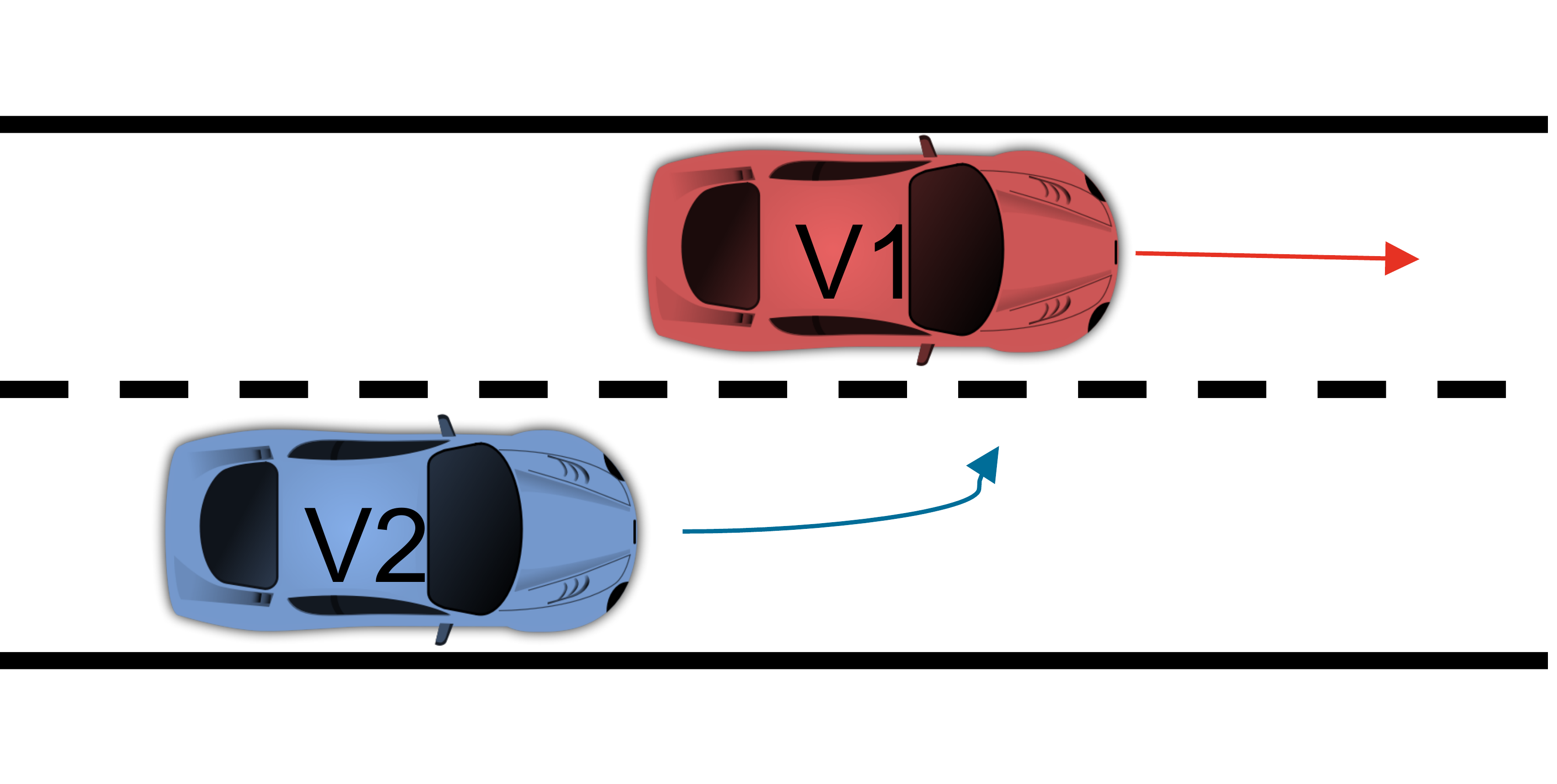}
         \caption{Beginning}
         \label{fig:lcB1}
     \end{subfigure}
     \hfill
     \begin{subfigure}[b]{0.45\linewidth}
         \centering
         \includegraphics[width=\textwidth]{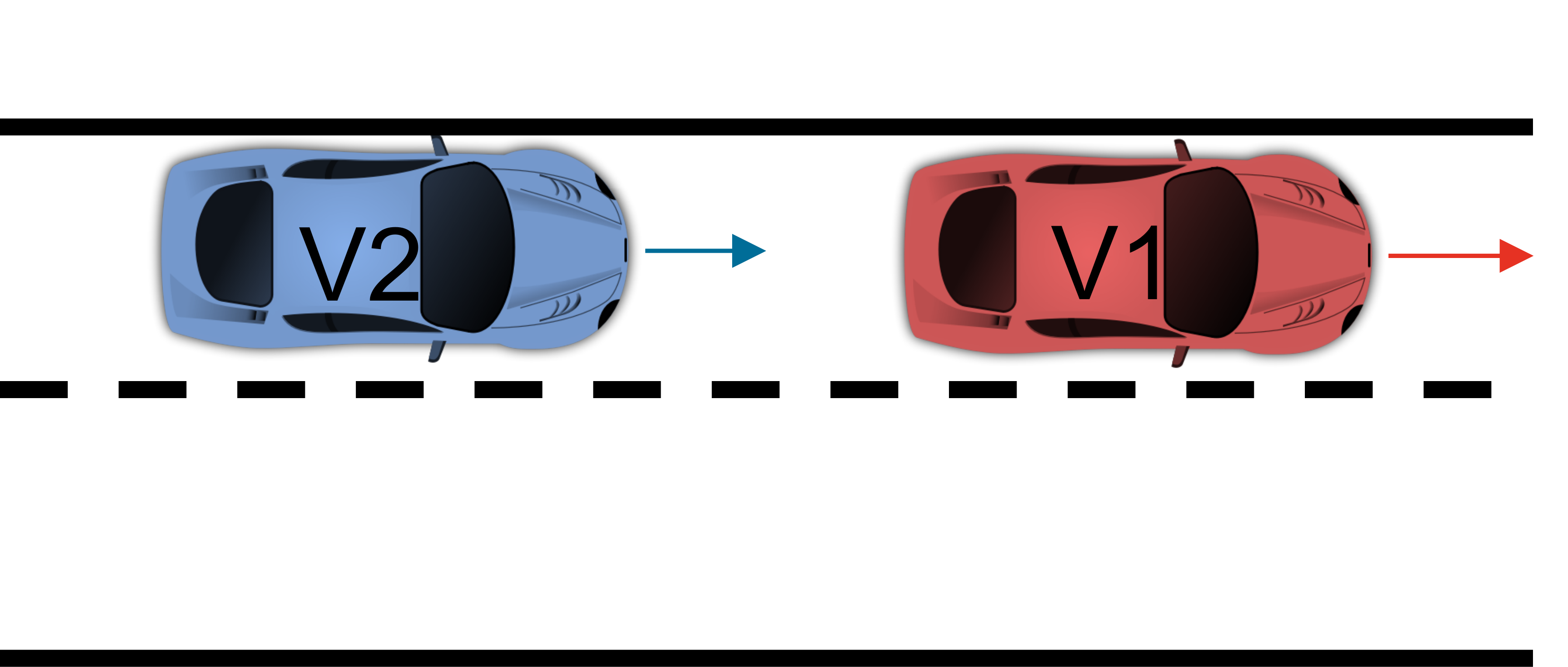}
         \caption{End}
         \label{fig:lcB2}
     \end{subfigure}
        \caption{2nd type of lane change. The V2 (blue) changes lane behind the vehicle V1 (red) to the left adjacent lane.}
        \label{fig:lc_typeB}
\end{figure}

\begin{enumerate}
		\item V1 changes lanes to its right, in front of V2, see Figure \ref{fig:lc_typeA}.
		\item V2 changes lanes to its left, behind V1, see Figure \ref{fig:lc_typeB}.
\end{enumerate}

In the first case 1), the driver in V1 estimates the distance or duration between the front part of the following vehicle V2. In the second case, the driver of V2 estimates the headway between V2 and V1.
Since no other vehicles are in the immediate vicinity, the only safety issue to take into account is a potential collision between the two involved vehicles.

\begin{figure}
	\centering
     \begin{subfigure}[b]{0.45\linewidth}
         \centering
         \includegraphics[width=\textwidth]{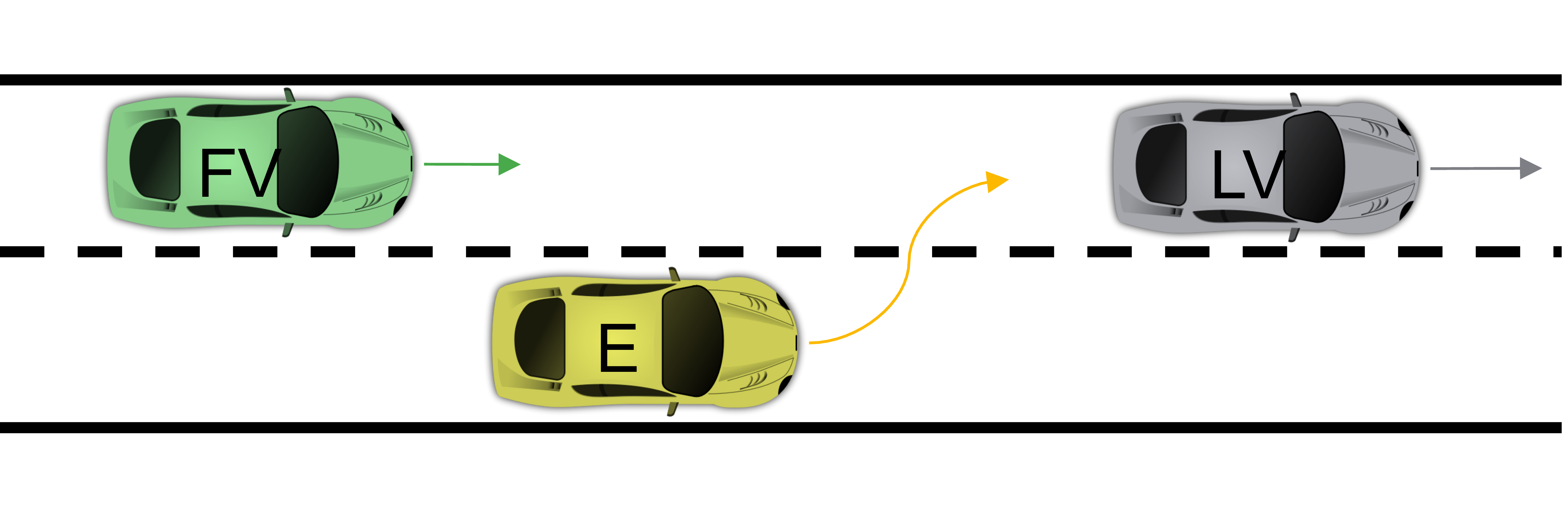}
         \caption{Beginning}
         \label{fig:lcC1}
     \end{subfigure}
     \hfill
     \begin{subfigure}[b]{0.45\linewidth}
         \centering
         \includegraphics[width=\textwidth]{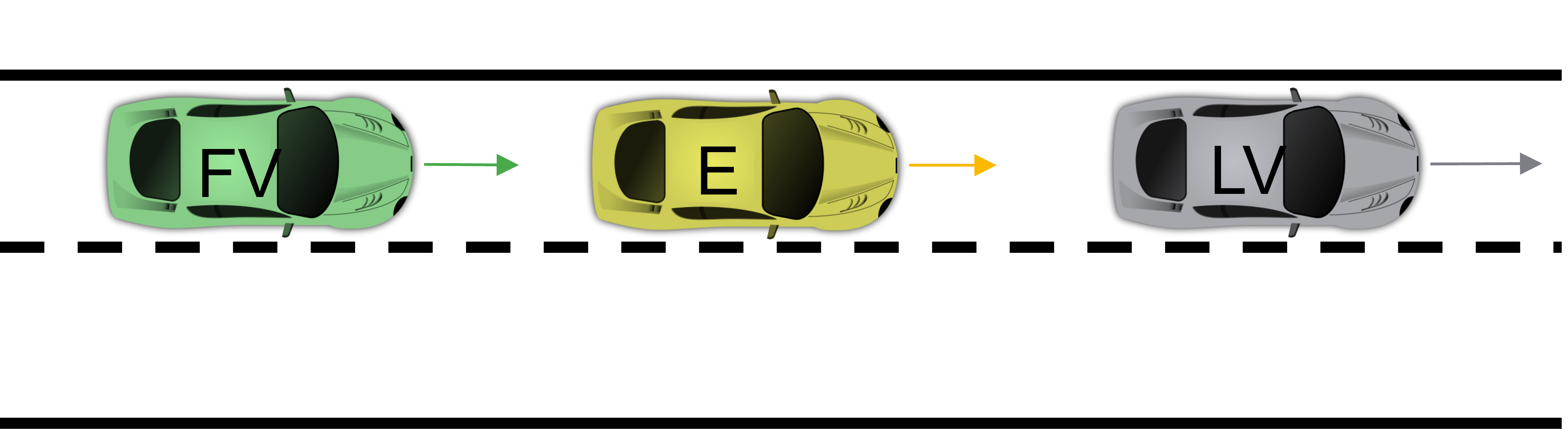}
         \caption{End}
         \label{fig:lcC2}
     \end{subfigure}
        \caption{Lane change in between two vehicles. The vehicle E (yellow) changes lane behind the leading vehicle LV (grey) and in front of the following vehicle FV (green). Note that the lane change can also occur to the right adjacent lane of vehicle E, thought the image is omitted here.}
        \label{fig:lc_typeC}
\end{figure}

We additionally considered a further scenario, in which a vehicle (E) changed lanes in between two others, a faster leading vehicle (LV) and a slower following vehicle (FV). In those situations, the driver of E estimates the ratio between both distances. However, this type of traffic situations is more difficult to categorize as subjectively safe, since the number of independent variables that affect safety is much higher than in the previously described scenario.
The traffic situation can be seen in Figure \ref{fig:lc_typeC}.

\subsection{HighD dataset}

\begin{figure}
	\centering
	\includegraphics[width=\linewidth]{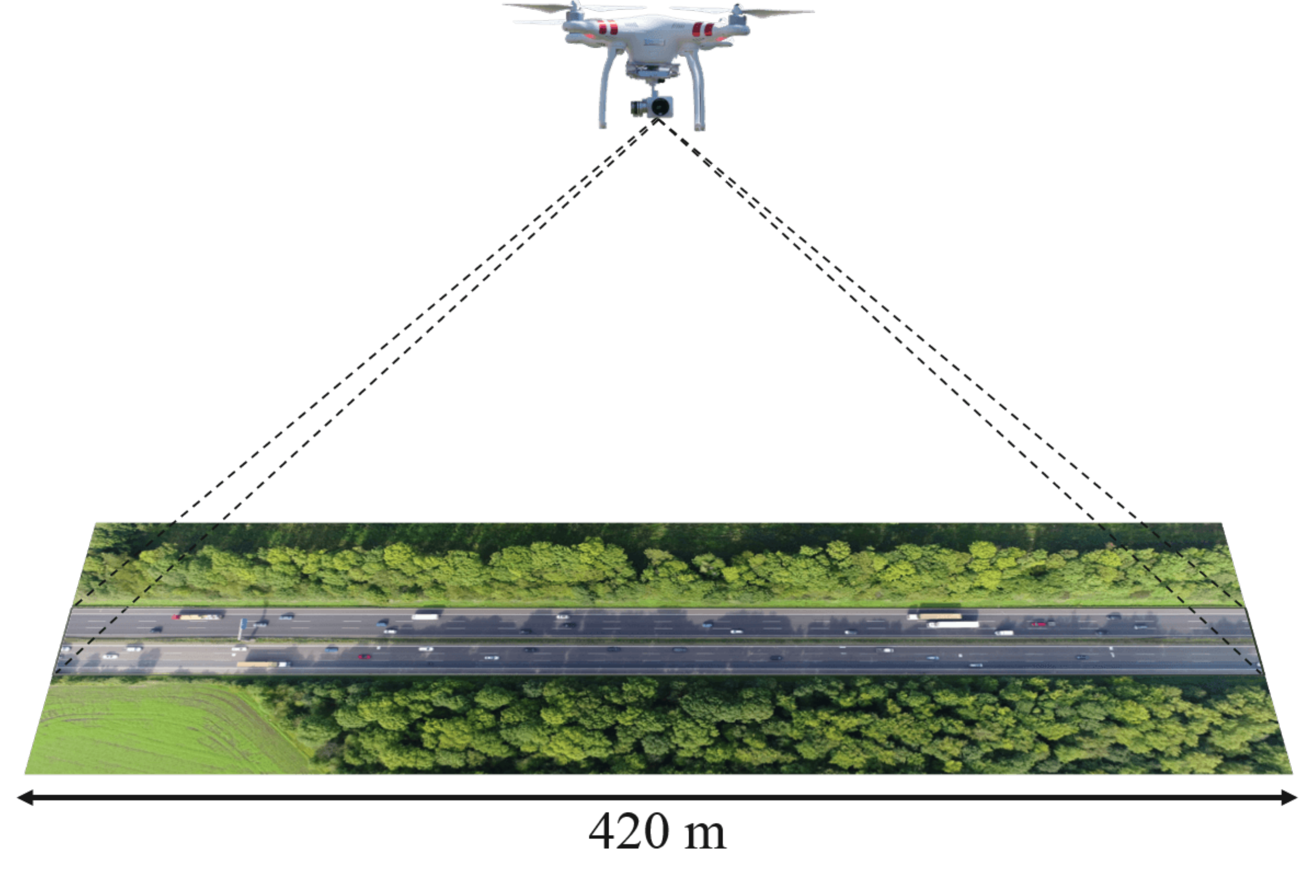}	
	\caption{Visualization of how the HighD dataset was recorded.}
	\label{fig:HighD}
\end{figure}

The scenarios studied in this paper were extracted from the HighD dataset \cite{highDdataset}, which includes trajectories of cars and trucks on German highways that were recorded by a drone over a total of 16.5 h. One of the recorded highway sections is shown in Figure \ref{fig:HighD}. Trucks were not considered in our scenario, thus any interaction with them was excluded. Other individual differences between vehicles which could have influenced the scenario, such as the warning lights on a vehicle or a student driver, could not be accessed within this dataset and thus were not taken into consideration. However, the rarity of those events should not have influenced the quantitative results.

Extracting interactions between only cars and excluding all situations with nearby surrounding traffic (where the Time Headway (TH) was below 2 s), resulted in a total of 1546 lane change situations for the first scenario, namely lane changing after overtaking. Other overtaking maneuvers such as the one shown in Figure \ref{fig:overtaking} were not analyzed due to the lack of data.

Of the 1546 lane changes, the FV changed lanes in 832 situations (Figure \ref{fig:lc_typeB}) whereas the LV changed lanes in 714 (Figure \ref{fig:lc_typeA}). The beginning of the lane changing maneuver was set when the acting vehicle began crossing the lane boundaries. At that moment in time, it is a reasonable assumption that the distance and the surrogate safety measures (see following section) would correspond to a safe situation for the acting vehicle. Nonetheless, due to the limited length of the recorded highway sections (around 420 m), it was not possible to exclude the existence of unknown factors influencing the trajectories.

For the second scenario the estimation of potentially unsafe situations was more challenging as a higher traffic density and therefore more variables affected the outcome. The 316 traffic situations were compared to the overtaking scenarios by filtering situations in which the Time Headway between FV and E and between E and LV were below 1 [s] at the beginning of the lane change, as defined previously. Larger TH indicated a lack of interaction between the participants and were thus ignored.

Most traffic situations in the second scenario corresponded exactly to Figure \ref{fig:lc_typeC} were E changed to its left adjacent lane. Changes to the right adjacent lane were typically coupled with a slower LV and faster FV and thus not considered in this paper.

\subsection{Safety Measures}

Since subjective safety depends on several factors such as experience, emotions, cognitive load, etc~\cite{olaverri2017road}, 
a quantifiable measurement based on Surrogate Safety Measures (SSMs) is more reliable to estimate traffic safety. In this case SSMs use distance, velocity etc., to assign a numerical value to a specific traffic situation. Traffic control techniques \cite{STCT_origin} use these values and empirically establish thresholds to determine whether traffic situations are safe or not and have been shown to successfully extrapolate traffic safety to accident rate.

Two SSMs were chosen for an objective safety assessment, Time Headway (TH) and Potential Index for Collision with Urgent Deceleration (PICUD) (see equations (1) and (2)). Both are well suited for lane-change and car-following situations, the former describes the time it takes the following vehicle (FV) to pass the position of the leading vehicle (LV) whereas the latter describes the distance between the two vehicles should both the LV and FV decelerate to a full stop. PICUD values less than or equal to 0 m characterize a traffic situation as potentially dangerous, for TH typical thresholds are around 1 [s] \cite{SSM}.

\begin{equation}
	TH = \frac{\Delta_x}{v_{FV}}
\end{equation}

\begin{equation}
	PICUD = \frac{v_{LV}^2-v_{FV}^2}{2\alpha}+\Delta_x -v_{FV}\Delta_t,
\end{equation}
where $v_{LV}$ and $v_{FV}$ are the velocity of the LV and FV respectively, $\Delta_x$ is the distance between them, $\alpha = 3.3 [m/s^2]$ is the sudden deceleration rate of both vehicles and $\Delta_t = 1[s]$ is the reaction time of the following vehicle. 

Other SSMs for car-following scenarios, such as Time To Collision (TTC) or Deceleration Rate To Avoid Crash (DRAC), were not relevant for this paper since V1 in the dataset was always faster than the V2 and thus the vehicles were not on a collision course.

\subsection{Statistical Methods}

To compare the two different lane change situations, a Kolmogorov-Smirnov 2 sample test was used. The non parametric test can be used to test whether the underlying distributions of empirical data differ. We used a 95\% confidence level to test if the underlying distributions were identical (null hypothesis).
Based on the results of the Kolmogorov-Smirnov test a Spearman R test was used to further investigate potential differences and measure the dependency between variables describing the traffic situation. As in the previous test, we also chose a 95\% confidence level to test whether an uncorrelated system could have Spearman coefficients at least as high as the test results of the dataset. Spearman was selected because the resulting correlation coefficients are not limited to linear ones and is also a non parametric test.

\section{Results}
\label{sec:results}

\begin{figure}
	\centering
	\includegraphics[width=\linewidth]{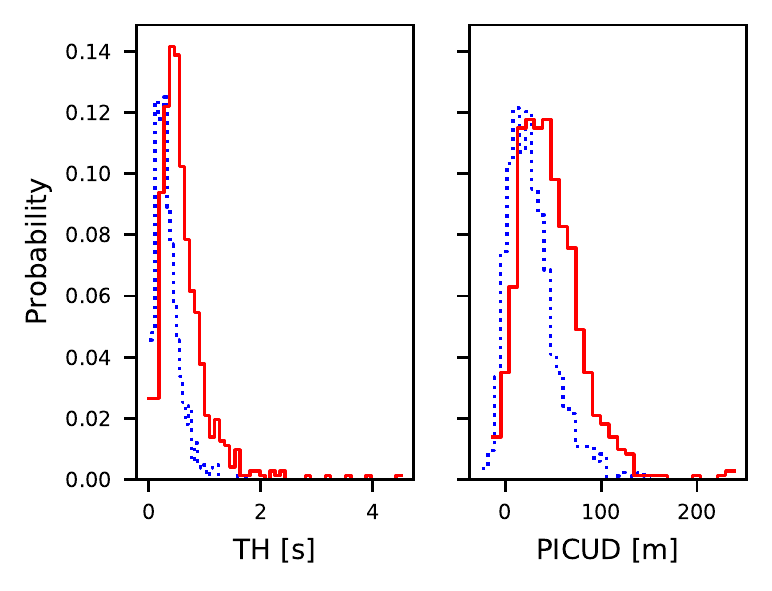}	
	\caption{Histogram of TH and PICUD for V1(red, -) and V2 (blue, $\cdot$) changing lane, normalized to probability. For both SSMs V2 changing lanes occurs at lower values, representing a riskier situation.}
	\label{fig:SSMs}
\end{figure}

For lane change after overtaking we calculated TH and PICUD as well as the distance and speed difference for V1 and V2 at the beginning of the lane change maneuver. The histogram, normalized to probability, of the two SSMs is shown in Figure \ref{fig:SSMs}, with the continuous red line representing lane changes of V1 and the dotted blue line the lane changes of V2. 

\begin{figure}
	\centering
	\includegraphics[width=\linewidth]{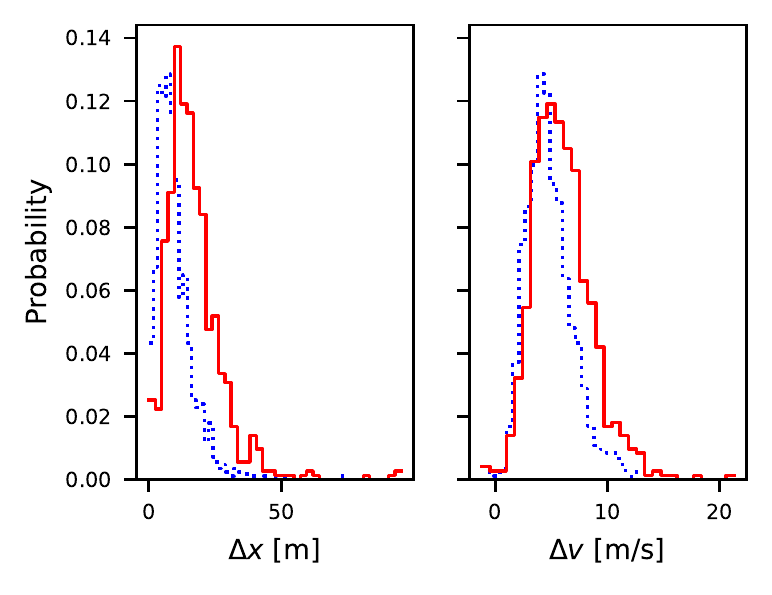}	
	\caption{Histogram of $\Delta x$, the distance between V1 and V2, and $\Delta v$,the speed difference between V1 and V2, for  V1(red, -) and V2 (blue, $\cdot$) changing lane, normalized to probability. The distance when V2 changes lane is significantly lower than for V1, though the speed difference shows a similar distribution in both lane change types.}
	\label{fig:Deltas}
\end{figure}

For both SSMs, but in particular TH, lane changes of V2 begin at lower values, thus showing a riskier situation between V1 and V2.
This is also visible in the histogram of the distance between the two vehicles, shown in Figure \ref{fig:Deltas}.

\begin{figure}
	\centering
	\includegraphics[width=\linewidth]{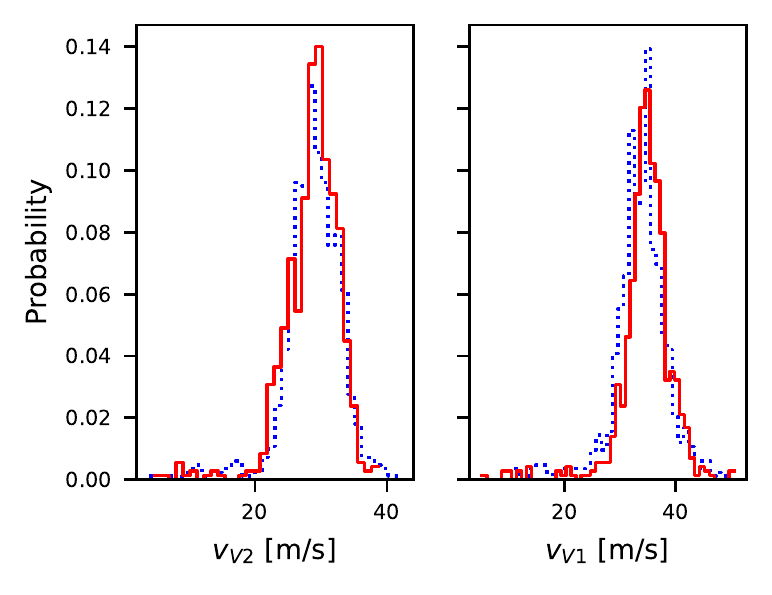}	
	\caption{Histogram of $v_{V1}$,velocity of V1, and $v_{V2}$, velocity of V2, for  V1(red, -) and V2 (blue, $\cdot$) changing lane, normalized to probability. In both lane change types the distributions appear to be similar.}
	\label{fig:vs}
\end{figure}

On the other hand, the velocities (Figure \ref{fig:Deltas}) and the absolute velocity of V1 and V2, shown in Figure \ref{fig:vs}, do not differ substantially between both types of lane change.

This can be seen in more detail in Table \ref{tab:Means}, where the means of the aforementioned measures and velocities are shown. The means of TH, PICUD and the distance differ substantially depending on the lane change type, whereas the other values, which are not strictly safety related, do not. Having removed external influences, these results indicate that there is a substantial difference in the traffic situations that the drivers in V1 and V2 perceive as safe.

\npdecimalsign{.}
\nprounddigits{3}
\begin{table}[h]
\centering
\caption{Means of the parameters at the beginning of either lane-change scenarios}
\label{tab:Means}
	\begin{tabular}{l| n{2}{5} | n{2}{5}}
	{Metric} & {V2 lane-change} & {V1 lane-change}\\ [0.5ex]
	\hline
	$\text{TH}$ [s]&$0.374562$&$0.616286$\\ 
	\hline
	$\text{PICUD}$ [m]&$30.774802$&$50.345132$\\
	\hline
	$\Delta x$ [m]&$10.733757$&$17.265354$\\
	\hline
	$\Delta v$ [m/s]&$5.061575$&$6.361181$\\
	\hline
	$v_{V2}$ [m/s]&$28.969829$&$28.878504$\\
	\hline
	$v_{V1}$ [m/s]&$34.031404$&$35.239685$\\
	\hline
	\end{tabular}
\end{table}
\npnoround

\npdecimalsign{.}
\nprounddigits{3}
\begin{table}[h]
\caption{Kolmogorov-Smirnov 2 sample test results. K is the direct result, P is the probability with which two samples of the two distributions would originate from the same.}
\label{tab:KS}
\begin{center}
	\begin{tabular}{l|n{2}{5}|n{2}{5}} 
	{Metric}& {K}& {P}\\ [0.5ex] 
	\hline
	$\text{TH}$&$0.3751322917238298$&$2.2638364452788246e-45$\\
	\hline
	$\text{PICUD}$&$0.3088629575333767$&$1.4303821051189241e-30$\\
	\hline
	$\Delta x$&$0.39091322446867727$&$2.2309475269835396e-49$\\	
	\hline
	$\Delta v$&$0.252001034815262$&$2.353195524099576e-20$\\
	\hline
	$v_{V2}$&$0.05244239920350583$&$0.2704151965967384$\\
	\hline
	$v_{V1}$&$0.1759813106092083$&$4.3581083186694514e-10$\\
	\hline
\end{tabular}
\end{center}
\end{table}
\npnoround

\npdecimalsign{.}
\nprounddigits{3}
\begin{table}[h]
\caption{Spearman R correlation for parameters at the beginning of the lane-change.}
\setlength\tabcolsep{1.5pt}
\label{tab:S_corr}
\begin{center}
	\begin{tabular}{l|n{2}{5}|n{2}{5}|n{2}{5}|n{2}{5}|n{2}{5}|n{2}{5}}
	{Metric}& $\text{TH}$& $\text{PICUD}$&$\Delta x$&$\Delta v$&$v_{V2}$&$v_{V1}$\\ [0.5ex] 
	\hline
	$\text{TH}$&$1.000000$&$0.607272$&$0.965602$&$0.397935$&$-0.171108$&$0.086920$\\
	\hline
	$\text{PICUD}$&$0.607272$&$1.000000$&$0.616929$&$0.945771$&$-0.087335$&$0.456357$\\
	\hline
	$\Delta x$&$0.965602$&$0.616929$&$1.000000$&$0.376432$&$0.031704$&$0.254970$\\	
	\hline
	$\Delta v$&$0.397935$&$0.945771$&$0.376432$&$1.000000$&$-0.195215$&$0.377811$\\
	\hline
	$v_{V2}$&$-0.171108$&$-0.087335$&$0.031704$&$-0.195215$&$1.000000$&$0.787800$\\
	\hline
	$v_{V1}$&$0.086920$&$0.456357$&$0.254970$&$0.377811$&$0.787800$&$1.000000$\\
	\hline
\end{tabular}
\end{center}
\end{table}
\npnoround

\npdecimalsign{.}
\nprounddigits{3}
\begin{table}[h]
\caption{Spearman R correlation significance test of parameters at the beginning of the lane-change. A value below 0.05 means the correlation is significant.}
\label{tab:S_sign}
\begin{center}
	\begin{tabular}{l|c|c|c|c|c|c|}
	{Metric}& $\text{TH}$& $\text{PICUD}$&$\Delta x$&$\Delta v$&$v_{V2}$&$v_{V1}$\\ [0.5ex] 
	\hline
	$\text{TH}$&$0$&$0$&$0$&$0$&$0$&$0.005$\\
	\hline
	$\text{PICUD}$&$0$&$0$&$0$&$0$&$0.005$&$0$\\
	\hline
	$\Delta x$&$0$&$0$&$0$&$0$&$0.308$&$0$\\	
	\hline
	$\Delta v$& $0$ & $0$ & $0$ & $0$ & $0$ & $0$ \\
	\hline
	$v_{V2}$ & $0$ & $0.005$ & $0.308$ & $0$ & $0$ & $0$\\
	\hline
	$v_{V1}$& $0.005$ & $0$ & $0$ & $0$ & $0$ & $0$ \\
	\hline
\end{tabular}
\end{center}
\end{table}
\npnoround

The results from the Kolmogorov-Smirnov 2 sample test (Table \ref{tab:KS}) also indicated a statistically significant difference for objective safety parameters between situations where V1 or V2 changed lanes. The initial traffic situation, i.e. distance and speed of the vehicles, also differed significantly, with the exception for the velocity of V2.
However, the correlation between $v_{V2}$ and objective safety metrics was both weak and partially not significant, as can be seen in Table \ref{tab:S_corr} and Table \ref{tab:S_sign}. In general it only seems to influence velocity of V1, necessary since V1 has overtaken V2, and thus the lack of difference according to the Kolmogorov-Smirnov 2 sample test does not indicate that the two lane-changing situations do not statistically differ significantly from each other.
Note that the correlation was calculated for all lane-change maneuvers, though the results do not change significantly when split up by type of lane-change.

Summarizing, the probability of the data originating from the same distribution was not statistically significant according to nearly all metrics, rejecting thus the previously defined null hypothesis. The only exception was the distribution of $v_{V2}$. In this case, the hypothesis was accepted.

Though finding the reason for the discrepancy of safety perception is not topic of this paper, it is necessary to mention that the distortion of the perception of distance might be due to technical reasons. A study \cite{mirror} of distance perception with convex mirrors (as most side mirrors are) showed how distance was indeed estimated incorrectly, though not to the extent the distance between V1 and V2 differed according to type of lane change.

For the second scenario, lane change in between two vehicles, we calculated the same safety measures. Furthermore, we used the ratio between the safety measure to LV and FV to check whether we obtained similar results.

\begin{figure}
	\centering
	\includegraphics[width=\linewidth]{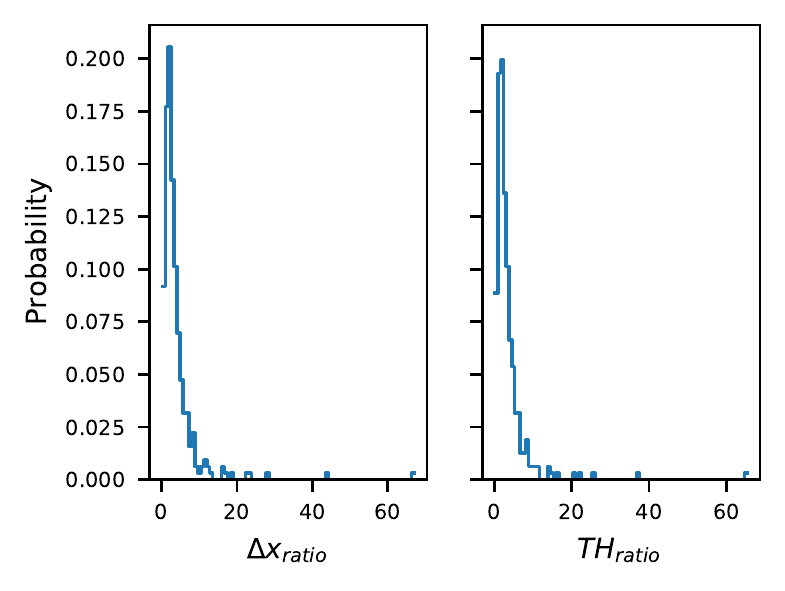}	
	\caption{Histogram of distance and TH ratios for the lane change in between two vehicles, normalized to probability.}
	\label{fig:ratios}
\end{figure}

The ratios were defined as depicted in (3) and (4).
\begin{equation}
	TH_{ratio} = \frac{TH_{FV}}{TH_{LV}},
\end{equation}
where $TH_{FV}$ is the TH between FV and E and $TH_{LV}$ is the TH between E and LV. And
\begin{equation}
	\Delta_{x_{ratio}} = \frac{\Delta_{x_{FV}}}{\Delta_{x_{LV}}},
\end{equation}
where $\Delta_{x_{FV}}$ is the distance between FV and E and $\Delta_{x_{LV}}$ is the distance between E and LV. The histogram of both ratios, normalized to probability, is shown in Figure \ref{fig:ratios}, with the mean values shown in Table(\ref{tab:ratios}). On average, and even when eliminating outliers (largest 1\%), the TH and distance between E and LV are significantly smaller than between FV and E. This coincides with the previous observations, where a larger safe distance was kept when changing lane in front of V2 than behind V1.

\npdecimalsign{.}
\nprounddigits{3}
\begin{table}[h]
\centering
\caption{Means of the ratios at the beginning the lane change for both all scenarios and removing outliers.}
\label{tab:ratios}
	\begin{tabular}{l| n{2}{2} |n{2}{2}}
	{Metric} & {Ratio (all)} &{Ratio (no outliers)}\\ [0.5ex]
	\hline
	$\text{TH}$ &$3.651610$&$3.217141$\\ 
	\hline
	$\Delta_{x}$&$4.006619$&$3.538249$\\
	\hline
	\end{tabular}
\end{table}
\npnoround

\begin{figure}
	\centering
	\includegraphics[width=0.9\linewidth]{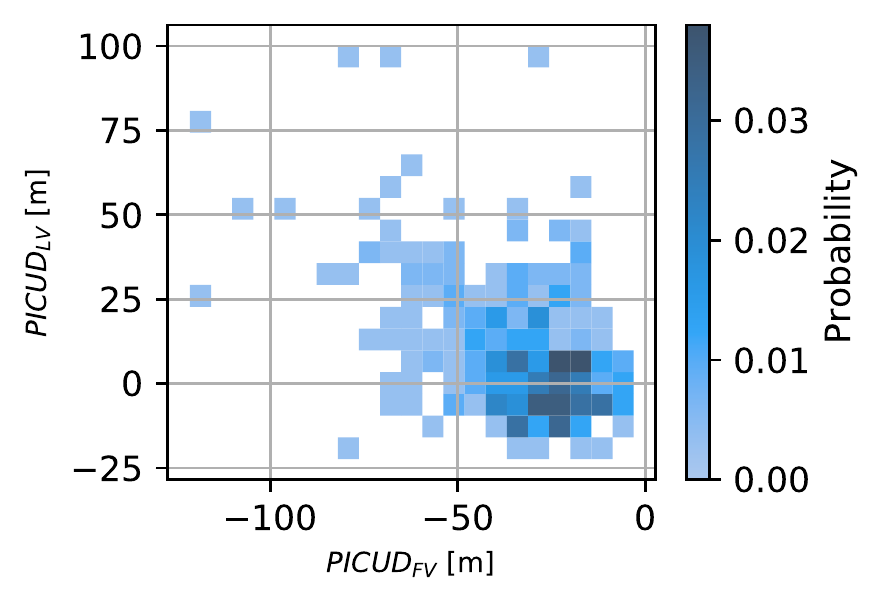}	
	\caption{2-dimensional histogram of the PICUD values between FV and E ($PICUD_{FV})$ and E and LV $(PICUD_{LV})$, normalized to probability.}
	\label{fig:PICUD2d}
\end{figure}

For PICUD a similar approach did not work, since it can be positive or negative, thus a 2-dimensional histogram was used, as shown in Figure \ref{fig:PICUD2d}. 

Interestingly, the values showed an opposite situation than TH or distance, and contradicted thus the results from the first scenario. According to PICUD the lane changes occurred in a safer traffic situation with respect to LV than FV. 

A large difference between both scenarios could be appreciated in the direction of the lane changes. For the lane change after overtaking, V1 always changed to its right adjacent lane whereas V2 always changed to its left lane.
The data from the dataset showed that for the scenario where the lane change occurred in between two vehicles, most lane changes occurred to the left adjacent lane. This was due to the selection of lane changes where the leading vehicle was faster than E and the following vehicle was slower. Obtaining the statistical significance of this difference was not possible due to a lack of data.
Also, the influence of surrounding vehicles in the second scenario was much higher than in the first one, in which the traffic was much less dense.

Therefore, while it is reasonable to claim that subjective safety has a road orientation component when compared to SSMs, the extent of it and the overall implications as well as the potential influence of a lateral metric require further analysis to start building a bridge between subjective safety and objective safety measures.

\section{CONCLUSION}
\label{sec:conclusion}

A difference in subjective safety could be determined in this paper by analyzing overtaking and lane-change situations in a highway scenario. With equal surrounding conditions a higher distance and safety thresholds were kept when safety had to be calculated with respect to the slower following vehicle than the faster leading vehicle. 

A further comparison with lane changes in between vehicles showed results which were partially in agreement with the initial results, though not fully. 
The direction of the lane change (right or left adjacent lane) could also be an influencing factor requiring future work and data analyses.

Other unknown factors might also contribute to the discrepancy in some of the results. Therefore, further research will be performed to achieve conclusions regarding whether or not an AV should mimic driver behavior.

\addtolength{\textheight}{-6cm}   





\section*{ACKNOWLEDGEMENT}


This work was supported by the Austrian Science Fund
(FWF), within the project ”Interaction of autonomous and manually-controlled vehicles (IAMCV)”, number P 34485-N.

\bibliographystyle{IEEEtran}
\bibliography{biblio}

\begin{thebibliography}{10}
\providecommand{\url}[1]{#1}
\csname url@samestyle\endcsname
\providecommand{\newblock}{\relax}
\providecommand{\bibinfo}[2]{#2}
\providecommand{\BIBentrySTDinterwordspacing}{\spaceskip=0pt\relax}
\providecommand{\BIBentryALTinterwordstretchfactor}{4}
\providecommand{\BIBentryALTinterwordspacing}{\spaceskip=\fontdimen2\font plus
\BIBentryALTinterwordstretchfactor\fontdimen3\font minus
  \fontdimen4\font\relax}
\providecommand{\BIBforeignlanguage}[2]{{%
\expandafter\ifx\csname l@#1\endcsname\relax
\typeout{** WARNING: IEEEtran.bst: No hyphenation pattern has been}%
\typeout{** loaded for the language `#1'. Using the pattern for}%
\typeout{** the default language instead.}%
\else
\language=\csname l@#1\endcsname
\fi
#2}}
\providecommand{\BIBdecl}{\relax}
\BIBdecl

\bibitem{olaverri2016autonomous}
C.~Olaverri-Monreal, ``Autonomous vehicles and smart mobility related
  technologies,'' \emph{Infocommunications Journal}, vol.~8, no.~2, pp. 17--24,
  2016.

\bibitem{olaverri2020promoting}
------, ``Promoting trust in self-driving vehicles,'' \emph{Nature
  Electronics}, vol.~3, no.~6, pp. 292--294, 2020.

\bibitem{subjective}
Z.~Li, X.~Zhou, X.~Wang, and Z.~Guo, ``Study on subjective and objective safety
  and application of expressway,'' \emph{Procedia - Social and Behavioral
  Sciences}, vol.~96, pp. 1622--1630, 11 2013.

\bibitem{smirnov2021game}
N.~Smirnov, Y.~Liu, A.~Validi, W.~Morales-Alvarez, and C.~Olaverri-Monreal, ``A
  game theory-based approach for modeling autonomous vehicle behavior in
  congested, urban lane-changing scenarios,'' \emph{Sensors}, vol.~21, no.~4,
  p. 1523, 2021.

\bibitem{validi2022metamodel}
A.~Validi, N.~Smirnov, and C.~Olaverri-Monreal, ``Metamodel-based simulation to
  assess platooning on battery energy consumption,'' \emph{Transportation
  Research Part D: Transport and Environment}, vol. 109, p. 103350, 2022.

\bibitem{Courteous}
L.~Sun, W.~Zhan, M.~Tomizuka, and A.~Dragan, ``Courteous autonomous cars,'' 10
  2018, pp. 663--670.

\bibitem{factorslanechange}
\BIBentryALTinterwordspacing
M.~Shawky, ``Factors affecting lane change crashes,'' \emph{IATSS Research},
  vol.~44, no.~2, pp. 155--161, 2020. [Online]. Available:
  \url{https://www.sciencedirect.com/science/article/pii/S0386111219300020}
\BIBentrySTDinterwordspacing

\bibitem{ESRA}
K.~S. S. N. S.~A. Furian, G., ``Subjective safety and risk perception,'' 2021.

\bibitem{SSM}
S.~M. Mahmud, L.~Ferreira, M.~Hoque, and A.~Hojati, ``Application of proximal
  surrogate indicators for safety evaluation: A review of recent developments
  and research needs,'' \emph{IATSS Research}, vol.~41, 03 2017.

\bibitem{DSSM}
S.~Tak, S.~Kim, D.~Lee, and H.~Yeo, ``A comparison analysis of surrogate safety
  measures with car-following perspectives for advanced driver assistance
  system,'' \emph{Journal of Advanced Transportation}, vol. 2018, 11 2018.

\bibitem{highDdataset}
R.~Krajewski, J.~Bock, L.~Kloeker, and L.~Eckstein, ``The highd dataset: A
  drone dataset of naturalistic vehicle trajectories on german highways for
  validation of highly automated driving systems,'' in \emph{2018 21st
  International Conference on Intelligent Transportation Systems (ITSC)}, 2018,
  pp. 2118--2125.

\bibitem{olaverri2017road}
C.~Olaverri-Monreal, ``Road safety: Human factors aspects of intelligent
  vehicle technologies,'' in \emph{Smart Cities, Green Technologies, and
  Intelligent Transport Systems}.\hskip 1em plus 0.5em minus 0.4em\relax
  Springer, 2017, pp. 318--332.

\bibitem{STCT_origin}
C.~Hydén and L.~Linderholm, ``The swedish traffic-conflicts technique,''
  \emph{International Calibration Study of Traffic Conflict Techniques}, pp.
  133--139, 01 1984.

\bibitem{mirror}
K.~Shimono and A.~Higashiyama, ``Perceived size and distance of virtual targets
  in convex mirrors,'' \emph{Perception}, vol.~29, pp. 33--33, 01 2000.

\end{thebibliography}
%
%
%
%
%
%

\end{document}